\pdfoutput=1

\documentclass[11pt]{article}

\usepackage[]{acl}

\newcommand{\anonymize}[2]{#2}

\usepackage{times}
\usepackage{latexsym}

\usepackage[T1]{fontenc}

\usepackage[utf8]{inputenc}

\usepackage{microtype}

\usepackage{amsmath}        
\usepackage{multirow}       
\usepackage{graphicx}
\usepackage{subcaption}     
\usepackage{vcell}          
\usepackage{amssymb}        
\usepackage{multicol}       
\usepackage{booktabs}       
\hypersetup{pdftitle={A New Aligned Simple German Corpus},
            pdfauthor={Vanessa Toborek, Moritz Busch, Malte Boßert, Christian Bauckhage, Pascal Welke},
            pdfkeywords={text simplification, sentence alignment, dataset, nlp},
            bookmarksnumbered=true
            }

\definecolor{darkblue}{rgb}{0.0, 0.0, 0.55}
\newcommand{\question}[2]{{\color{darkblue}{\textbf{#1} #2}}}
\newcommand{\answer}[1]{#1 \medskip}

\DeclareMathOperator{\tfidf}{tf-idf}

\DeclareMathOperator{\sentsim}{sim}
\DeclareMathOperator{\asym}{asym}
\DeclareMathOperator{\emb}{emb}
\DeclareMathOperator{\cossim}{cos_{sim}}
\newcommand{\mylist}[1]{\left[ #1 \right]}
\newcommand{\abs}[1]{\left \lvert #1 \right \rvert}
\newcommand{\norm}[1]{\left \lVert #1 \right \rVert}
\newcommand{\R}{\mathbb{R}}

\title{A New Aligned Simple German Corpus}

\author{
\textbf{Vanessa Toborek$^1$}  \textbf{Moritz Busch$^1$}  \textbf{Malte Boßert$^1$}  \textbf{Christian Bauckhage$^{1,2}$}  
\vspace{0.2cm}
\textbf{Pascal Welke$^{1,3}$} \\
$^1$University of Bonn, $^2$Fraunhofer IAIS, $^3$TU Wien \\
\texttt{\{toborek, welke\}@cs.uni-bonn.de}\\
\texttt{\{busch, mbossert\}@uni-bonn.de}\\
\texttt{christian.bauckhage@iais.fraunhofer.de}}
  
\begin{document}
\maketitle
\begin{abstract}
``Leichte Sprache'', the German counterpart to Simple English, is a regulated language aiming to facilitate complex written language that would otherwise stay inaccessible to different groups of people.
We present a new sentence-aligned monolingual corpus for Simple German -- German. 
It contains multiple document-aligned sources which we have aligned using automatic sentence-alignment methods.
We evaluate our alignments based on a manually labelled subset of aligned documents.
The quality of our sentence alignments, as measured by the F1-score,  surpasses previous work.
We publish the dataset under CC BY-SA and the accompanying code under MIT license.
\end{abstract}

\section{Introduction}
\label{sec:intro}

Text in simple language benefits language learners, people with learning difficulties, and children that tend to have a hard time understanding original and especially formal texts due to grammar and vocabulary.
Text simplification describes the problem of generating a simplified version of a given text while conveying the same matter \citep{Siddharthan2014}. This involves the reduction of lexical and syntactic complexity by various operations like deletion, rewording, insertion, and reordering \citep{Saggion2017}.
Text simplification can further entail additional explanations for difficult concepts and a structured layout \citep{Siddharthan2014}.

To make language more inclusive, guidelines for simple versions of languages exist. 
In English, most notably, \citet{ogden1932}  introduced ``Basic English''.
In German there are two prevalent kinds of simple language: ``Einfache Sprache'' (ES) and ``Leichte Sprache'' (LS), both roughly translating to \emph{easy language} \citep{Maass2020}. 
LS has strict rules, including the removal of subordinate clauses, the insertion of paragraphs after each sentence and the separation of compound nouns with hyphens. 
ES is less restrictive and does not have a specific set of rules; instead, translators can work more liberally. 
However, the goal of both approaches is to improve the language's accessibility.

There exists work on rule-based approaches for text simplification in German \cite{Suter2016}, but the problem of text simplification can also be defined as a monolingual translation task. Then, the availability of data becomes a prerequisite in order to apply statistical machine learning models to it.
Especially sentence-aligned text constitutes the backbone of neural machine translation.
To the best of our knowledge, only the work of \citet{klaper2013} presents a parallel sentence-aligned corpus in German created from public web data.
Our work addresses the lack of data for text simplification in German and thus creates an aligned corpus of easy language and corresponding German texts. 
As there is no German equivalent to the Simple English Wikipedia, which provides cross-lingual references between Simple English and English articles, we had to rely on multiple sources offering a small number of articles in German as well as in some simplified version of it. 
Our corpus consists of articles in ``Leichte Sprache'' from seven websites and ``Einfache Sprache'' from one extensive website. In the following, we will always talk about Simple German whenever the distinction between those two forms of simplification is not relevant.

Following the description of our dataset and its collection process, we present the results of a comparison of different sentence-alignment methods.
Then, we select the best approach and obtain a sentence-aligned dataset that can potentially be extended by crawling further websites. See \autoref{fig:examples} to see examples of our sentence alignments.
Finally, we discuss the limitations of our dataset and future research.
We share our code to build the dataset on GitHub\footnote{\anonymize{Anonymized.}{\href{https://github.com/mlai-bonn/Simple-German-Corpus}{https://github.com/mlai-bonn/Simple-German-Corpus}}}.
The repository contains a list of URLs and scripts to reproduce the dataset by crawling the archived websites, parsing the text and aligning the sentences.
We provide the fully prepared dataset upon request.

\begin{figure*}
    \small
    \centering
        \begin{tabular}{p{1cm}p{7cm}p{6.6cm}}
        \toprule
        \multirow{2}{*}{\begin{tabular}[c]{@{}l@{}}Not \\ Aligned\end{tabular}} 
                & {[}AS{]} Sämtliche Bälle müssen kugelförmig sein. & All balls must be spherical in shape. \\
                & {[}LS{]} Die Fahnen sollen bunt sein. & The flags should be colorful. \\
        \midrule
        \multirow{2}{*}{\begin{tabular}[c]{@{}l@{}}Partially \\ Aligned\end{tabular}} 
                & {[}AS{]} Diverse öffentliche Verkehrsmittel bieten eine optimale Anbindung an die Hamburger Innenstadt, die Autobahn sowie den Flughafen. & Various means of public transport offer an optimal connection to Hamburg's city center, the highway as well as the airport. \\
                & {[}LS{]} Die fahren bis zur Autobahn und zum Flughafen. & They go all the way to the highway and the airport. \\
        \midrule
        \multirow{2}{*}{Aligned}           
                & {[}AS{]} Bei Milch ist, falls es sich nicht um Kuhmilch handelt, die Tierart des Ursprungs anzugeben. & For milk, if it is not cow's milk, indicate the animal species of origin. \\
                & {[}LS{]} Manchmal ist die Milch nicht von einer Kuh. Dann muss man sagen von welchem Tier die Milch ist. & Sometimes the milk is not from a cow. Then you have to say which animal the milk is from. \\
        \bottomrule
        \end{tabular}
    \caption{Example sentence pairs aligned between Simple German [LS] and German [AS] and their translations. Examples show successfully, partially and wrongly aligned sentences.}
    \label{fig:examples}
\end{figure*}
\section{Related Work}
\label{sec:related}

There are various classification systems for language with different aims.
The European Council has defined six proficiency levels A1 to C2 based on the competencies of language learners and applicable to multiple languages \citep{cefr}.
Yet, these are mainly intended to evaluate learners, not texts.
For English, the \href{www.lexile.com}{Lexile
scale} gives scores on reading proficiency, as well as text complexity, but has been criticized as carrying little qualitative meaning \citep{commoncorestatestandardsEnglishLanguageArts}.
A particularly early attempt at a ``simplified'', controlled English language is Basic English \citep{ogden1932}.
It is a subset of (rules and words of) English and aims at being easy to learn without restricting sentence length, complexity of content, or implicit context.
As a result, even ``easy'' texts, as measured on one of the above scales, may fall short in comprehensibility and accessibility. 
We focus on German texts which follow specifically designed rules that aim at being more inclusive to certain target groups. 
LS (Simple German) is designed for people with cognitive disabilities \citep{Maass2020, maassLeichteSpracheRegelbuch, netzwerkleichtespracheRegelnFuerLeichte}.
ES (Plain German) targets the dissemination of expert contents to lay people and is less comprehensible (and hence less inclusive), but more acceptable to larger audiences \citep{Maass2020}. 

There are some sources of monolingual parallel corpora for different languages.
English -- simplified English corpora have been created, e.g. from the Simple English Wikipedia (which does not adhere to any fixed simplification standard) \citep{coster2011, hwang2015, jiang2020, zhu2010}.
Using aligned articles from Wikipedia has been criticized, as (i) simple Wikipedia contains many complex sentences and (ii) sentence alignments are improbable, as the articles are often independently written \citep{xu2015problems}.
Hence, an alternative corpus of five difficulty levels targeted at children at different reading levels has been proposed \citep{xu2015problems, jiang2020}. 
Spanish \citep{bott2011}, Danish \citep{klerke2012}, and Italian \citep{brunato2016} corpora exist as well.

When narrowing the research field down to the German language, only a few resources remain.
\citet{klaper2013} crawl five websites that provide a total of 256 parallel German and Simple German articles, spanning various topics.
They provide sentence level alignments, and thus their result is the most similar dataset to ours that currently exists.
They use a sentence alignment algorithm based on dynamic programming with prior paragraph alignment based on bag-of-word cosine similarities and report for their alignments an F1-score of 0.085 on the ground truth.
\citet{sauberli2020} introduce two sentence-aligned corpora gathered from the Austrian Press Agency and from capito. Here, the authors align the sentences of the original texts with their corresponding translation in level A1 and B1 of the Common European Framework of Reference for Languages \citep{cefr}. 
The resulting simplifications are very different to the simplifications according to the rules of LS.
\citet{rios2021} extend this dataset by adding articles from a Swiss news outlet which publishes ``simplified'' summaries alongside its content which, however, do not adhere to any simplification standard.
Here, sentence-level alignments are not provided.
\citet{battisti2020} compile a corpus for Simple German that mostly consists of unaligned Simple German articles and 378 parallel article pairs, but without sentence-alignments.
\citet{aumiller2022} present an extensive document-aligned corpus by using the German children encyclopedia \href{https://klexikon.zum.de/}{``Klexikon''}. 
The authors align the documents by choosing corresponding articles from Wikipedia, making it unlikely that specific sentences can be matched. 
As republishing may lead to legal ramifications, only the Klexikon dataset is publicly available.
Overall, current German language text simplification datasets are rare, small, usually not publicly available, and typically not focused on inclusive Simple German.
\section{Dataset Description}
\label{sec:description}

As discussed, there are very few datasets tailored towards text simplification. 
Our work addresses this lack of data for Simple German. 
Problems besides text simplification like automatic accessibility assessment, text summarization, and even curriculum learning would benefit from that data.


\label{sec:composition}



We present a corpus consisting of 712 German and 708 corresponding Simple German articles from eight web sources spanning different topics. They were collected from websites maintaining parallel versions of the same article in German and Simple German. 
We made sure to only use freely available articles. \autoref{tab:sources} in the appendix provides an overview of all websites with a brief description of their content.
Further, through the proposed automatic sentence alignment, we obtain a collection of about 10\,304 matched German and Simple German sentences. We will assess the quality of the sentence alignment in \autoref{sec:manual_evaluation}. 

\begin{table*}
    \centering
    
\begin{tabular}{crrrrrrrrrr} 
    \toprule
    \multicolumn{1}{l}{} & \multicolumn{5}{c}{Simple German} & \multicolumn{5}{c}{German} \\ 
    \cmidrule(lr){2-6} \cmidrule(lr){7-11}
    source & \multicolumn{1}{c}{a} & \multicolumn{1}{c}{t} & \multicolumn{1}{c}{s/a} & \multicolumn{1}{c}{t/s} & \multicolumn{1}{c}{w/a}& \multicolumn{1}{c}{a} & \multicolumn{1}{c}{t} & \multicolumn{1}{c}{s/a} & \multicolumn{1}{c}{t/s} & \multicolumn{1}{c}{w/a}  \\ 
    \midrule
    apo & 168 & 94\,808 & 77.7 & 8.3 & 249.2 & 166 & 187\,427 & 78.8 & 16.4 & 543.3 \\
    beb & 21 & 5\,490 & 31.0 & 9.8 & 141.2 & 21 & 8\,131 & 23.5 & 18.7 & 216.3 \\
    bra & 47 & 9\,634 & 28.2 & 8.5 & 110.1 & 47 & 9\,728 & 13.3 & 18.9 & 142.7 \\
    lmt & 45 & 6\,946 & 20.0 & 9.2 & 99.9 & 45 & 9\,023 & 16.5 & 14.3 & 132.2 \\
    mdr & 322 & 53\,277 & 21.3 & 9.0 & 93.1 & 322 & 126\,191 & 29.8 & 15.1 & 235.4 \\
    soz & 15 & 5\,122 & 43.8 & 9.0 & 174.2 & 15 & 11\,790 & 61.0 & 14.8 & 437.7 \\
    koe & 82 & 66\,892 & 103.4 & 9.2 & 293.3 & 82 & 44\,310 & 42.4 & 14.3 & 265.8 \\
    taz & 8 & 7\,924 & 70.3 & 9.4 & 273.7 & 14 & 8\,171 & 41.1 & 16.6 & 336.6 \\ 
    \midrule
    total & 708 & 250\,093 & 49.5 & 9.1 & 179.3 & 712 & 404\,771 & 38.3 & 16.1 & 288.8 \\
    \bottomrule
\end{tabular}

    \caption{Statistics of the article-aligned corpus. (a) articles and (t) tokens per source; all further values are reported as averages: (s/a) sentences per article, (t/s) tokens per sentence, and (w/a) number of different words per article.}
    \label{tab:properties}
\end{table*}

Table \ref{tab:properties} shows statistics of the crawled and parsed articles.
In general, Simple German articles tend to be significantly shorter in the average number of words per article, while the number of sentences is  higher in Simple German than in German articles. This may be due to the fact that long sentences in German are split into multiple shorter sentences in Simple German. 
This motivates an $n:1$ matching between Simple German and German sentences.
\section{Dataset Construction}
\label{sec:construction}
We now describe the process of data acquisition from the selection of the online sources over the crawling of the websites to the parsing of the text. To be transparent, we point out the problems and pitfalls that we experienced during the process.

\paragraph{Crawling} 
Starting point for the construction of the dataset was a set of websites. \autoref{tab:sources} shows the websites that we used. These websites are publicly available, offer parallel articles in  German and Simple German, and cover a range of different topics. 
Many websites offer content in simple language, but few offer the same content parallel in German and in Simple German.
Hence, we ignored websites only in simple language. Due to its prevalence, most of the articles in our dataset are written in LS, but we also included one website in ES to increase the overall vocabulary size. In general, the data collection was limited by the availability of suitable and accessible data.

First, we identified a starting point for each website that offered an overview of all Simple German articles. 
Then, we created a crawling template for each website using the python library \href{https://www.crummy.com/software/BeautifulSoup/}{BeautifulSoup4}. The crawler always started from the articles in Simple German.
We first download the entire article webpages and later on parsed the text from the raw \texttt{html}-files. This process allows to return to the raw data to support unanticipated future uses. 

\paragraph{Parsing} We have ignored images, \texttt{html}-tags, and corresponding text metadata (e.g. bold writing, paragraph borders) for each article. In contrast to \citet{aumiller2022}, where enumerations are removed since they may only contain single words or grammatically incorrect sentences, we decided to transform them into comma-separated text. Enumerations are frequently used in Simple German articles, and we argue that they may contain major parts of information.

The most common challenge during crawling was an inconsistency in text location within a website, i.e. the structure of the \texttt{html}-boxes enclosing the main content.
Simply extracting by \texttt{<p>}-tag was not sufficient, as these regularly contained useless footer information.
As only the main text was the targeted resource, the crawler's implementation needed to be unspecific enough to account for these deviations, but specific enough not to crawl any redundant or irrelevant text.

Another problem was the way in which the German articles and their corresponding translations in Simple German were linked.
The \emph{mdr}, a state-funded public news organization, often showed inconsistent linking between articles.
Here one might expect a strict structure disallowing differences.
However, the links were sometimes encapsulated within \texttt{href}, sometimes given as plain text or not at all.
The referenced German article could even be a video, rendering both articles useless for our corpus. 
We discarded Simple German articles whenever the original  German source was unusable, i.e. unlocatable or in video format. 

The result of the data acquisition as described above is a dataset of articles in German with their corresponding articles in Simple German. 

\section{Sentence Alignment}
\label{sec:alignment}

In the following section we compare different similarity measures and matching algorithms used to reach sentence-level alignment. 
We describe an article $A$ as a list of sentences, i.e. $A = \mylist{s_1, \ldots, s_n}$. We define $A^S$ and $A^C$ as the simple and complex versions of the same article with $\abs{A^S} = n$ and $\abs{A^C} = m$. We consider a variant of the sentence alignment problem that receives two lists of sentences $A^S$ and $A^C$ and produces a list of pairs $\mylist{(s_i^S, s_j^C)}$ such that, with relative certainty, $s_i^S$ is a (partial) simple version of the complex sentence $s_j^C$. 
We will approach this task in three steps: 

First (Sec.~\ref{sec:preprocessing}), we transform the raw texts obtained in Section~\ref{sec:construction} into lists of sentences and do some light pre-processing.
Next, we compute sentence similarity scores (Sec.~\ref{sec:sim_measures}) for pairs of sentences from the aligned articles.
Finally, a sentence matching algorithm (Sec.~\ref{sec:matching}) takes the sentence lists and the respective inter-sentences similarities to calculate the most probable alignment. 

\subsection{Text Pre-processing}
\label{sec:preprocessing}

We apply a number of pre-processing steps to facilitate the sentence matching.
The sentence borders are identified using spaCy \citep{spacy}.
We neither apply lemmatization to the words nor do we remove stop words. 
All punctuation, including hyphens between compound nouns in Simple German, is removed. This pre-processing does not affect the final corpus. 

\textbf{Lowercase} letters are used for TF-IDF based similarity measures to decrease the vocabulary size. 
For similarity measures based on word vectors we apply no conversion:
The precomputed word vectors differ between lowercase and uppercase letters, e.g. ``essen'' (to eat) and ``Essen'' (food) or might not exist for their lowercase version.

\textbf{Gender-conscious suffixes} are removed. We are referring to word endings used in inclusive language to address female as well as other genders, not to endings that transform male nouns into their female form. In German, the female version of a word is often formed by appending ``-in'' (singular) or ``-innen'' (plural) to the end of the word, e.g. ``der Pilot'' (the male pilot) and ``die Pilotin'' (the female pilot). Traditionally, when talking about a group of people of unspecified gender, the male version was used. However, in order to include both men and women as well as other genders, different endings are preferred. The most popular ones are using an uppercase I (``PilotIn''), a colon (``Pilot:in''), an asterisk (``Pilot*in'') or an underscore (``Pilot\_in''). We remove these endings to make sentence matching easier. Such endings are commonly not included in Simple German texts. 

\subsection{Similarity Measures}
\label{sec:sim_measures}

After obtaining pre-processed lists of sentences $A^S$ and $A^C$, we compute similarities between any two sentences $s_i^S \in A^S$ and $s_j^C \in A^C$. A sentence can be described either as a list of words $s_i^S = \mylist{w_1^S, \ldots, w_l^S}$ or as a list of characters $s_i^S = \mylist{c_1^S, \ldots, c_k^S}$. 
In total, we have compared eight different similarity measures. Two of the measures are based on TF-IDF, the other six rely on word or sentence embeddings. We have decided to use the pre-trained fastText \cite{bojanowski2016} embeddings provided by spaCy's \href{https://spacy.io/models/de\#de_core_news_lg}{\texttt{d\_core\_news\_lg}} 
pipeline and the pre-trained \href{https://www.sbert.net/}{\texttt{distiluse-base-multilingual- cased-v1}} model for sentence embeddings provided by \citet{reimers2019}. 


\paragraph{TF-IDF based similarity measures} Both similarity measures calculate the cosine similarity $\cossim$ between two sentence vectors. We use the \emph{bag of word similarity} \citep{paetzold2017} that represents each sentence as a bag of word vector, weighted by calculating for each $w \in s_i$ the respective TF-IDF value. 
The \emph{character 4-gram similarity} \citep{stajner2018} works analogously, but uses character n-grams instead. 
We choose $n=4$. For further details see \autoref{appendix:ext_similarity_measures}.

\paragraph{Embedding based similarity measures} Using the pre-calculated word embeddings, the \emph{cosine similarity} calculates the angle between the average of each sentence's word vectors \citep{stajner2018, mikolov2013}.
The \emph{average similarity} \citep{kajiwara2016} calculates the average cosine similarity between all word pairs in a given pair $(A^S, A^C)$ using the embedding vector $\emb(w)$ of each word $w$.
In contrast, the \emph{Continuous Word Alignment-based Similarity Analysis} (CWASA) \citep{franco-salvador2015, stajner2018} does not average the embedding vectors. Instead, it finds the best matches for each word in $s^S$ and in $s^C$ with $\cossim \geq 0$. Then, the average cosine similarity is calculated between the best matches.
Likewise, the \emph{maximum similarity} \citep{kajiwara2016} calculates best matches for the words in both sentences. In contrast to CWASA, only the maximum similarity for each word in a sentence is considered.
Further, we implement the \emph{bipartite similarity} \citep{kajiwara2016} that calculates a maximum matching on the weighted bipartite graph induced by the lists of simple and complex words. Edges between word pairs are weighted with the word-to-word cosine similarity. The method returns the average value of the edge weights in the maximum matching.
The size of the maximum matching is bounded by the size of the smaller sentence.
Finally, we implement the \emph{SBERT} similarity by using a pre-trained multilingual SBERT model \citep{reimers2019, yang2020}. We calculate the cosine similarity on the contextualized sentence embeddings, cf. \autoref{appendix:ext_similarity_measures}.

\subsection{Matching Algorithms}
\label{sec:matching}

The previously presented methods are used to compute sentence similarity values for sentence pairs. Using these values, the sentence matching algorithm determines which sentences are actual matches, i.e. translations. For the two articles $\abs{A^S} = n$ and $\abs{A^C} = m$, the matrix $M \in \R^{n \times m}$ contains the sentence similarity measure for the sentences $s_i^S$ and $s_j^C$ in entry $M_{ij}$.
The goal is an $n:1$ matching of multiple Simple German sentences to one German sentence, but not vice versa. We explain the reasoning for this in Section~\ref{sec:composition}.

We compare two matching methods presented by \citet{stajner2018}. The first one is the most similar text algorithm (MST) which takes $M$ and matches each $s_i^S \in A^S$ with its most similar sentence in $A^C$.
The second method is the MST with Longest Increasing Sequence (MST-LIS). It is based on the assumption that the order of information is the same in both articles. 
It first uses MST and from this, only those matches appearing successively in the longest sequence are kept. All simple sentences not contained in that sequence are included in a set of unmatched sentences.
Let $(s_i^S, s_k^C), (s_j^S, s_l^C)$ be two matches in the longest sequence and $i < j \Rightarrow k \leq l$. Then, for all unmatched sentences $s_m^S$ with $i < m < j$, a matching $s^C$ will be looked for between indices $k$ and $l$. This is done iteratively for all sentences between $s_i^S$ and $s_j^S$. Corresponding matches cannot violate the original order in the Simple German article. 

We introduce a threshold that defines a minimum similarity value for all matched sentences. Simple sentences without any corresponding complex sentence will likely not be matched at all, as they are expected to have a similarity lower than the threshold to all other sentences.
Instead of picking a fixed value threshold as in \citet{paetzold2017}, we pick a variable threshold to consider that every similarity method deals with values in different ranges.
The threshold is set to $\mu(M) + k \cdot \sigma(M)$ with $\mu$ and $\sigma$ describing the mean of all sentence pair similarities and their standard deviation, respectively.
\section{Evaluation}
\label{sec:evaluation}

\begin{figure}
    \centering
    \includegraphics[width=0.9\columnwidth]{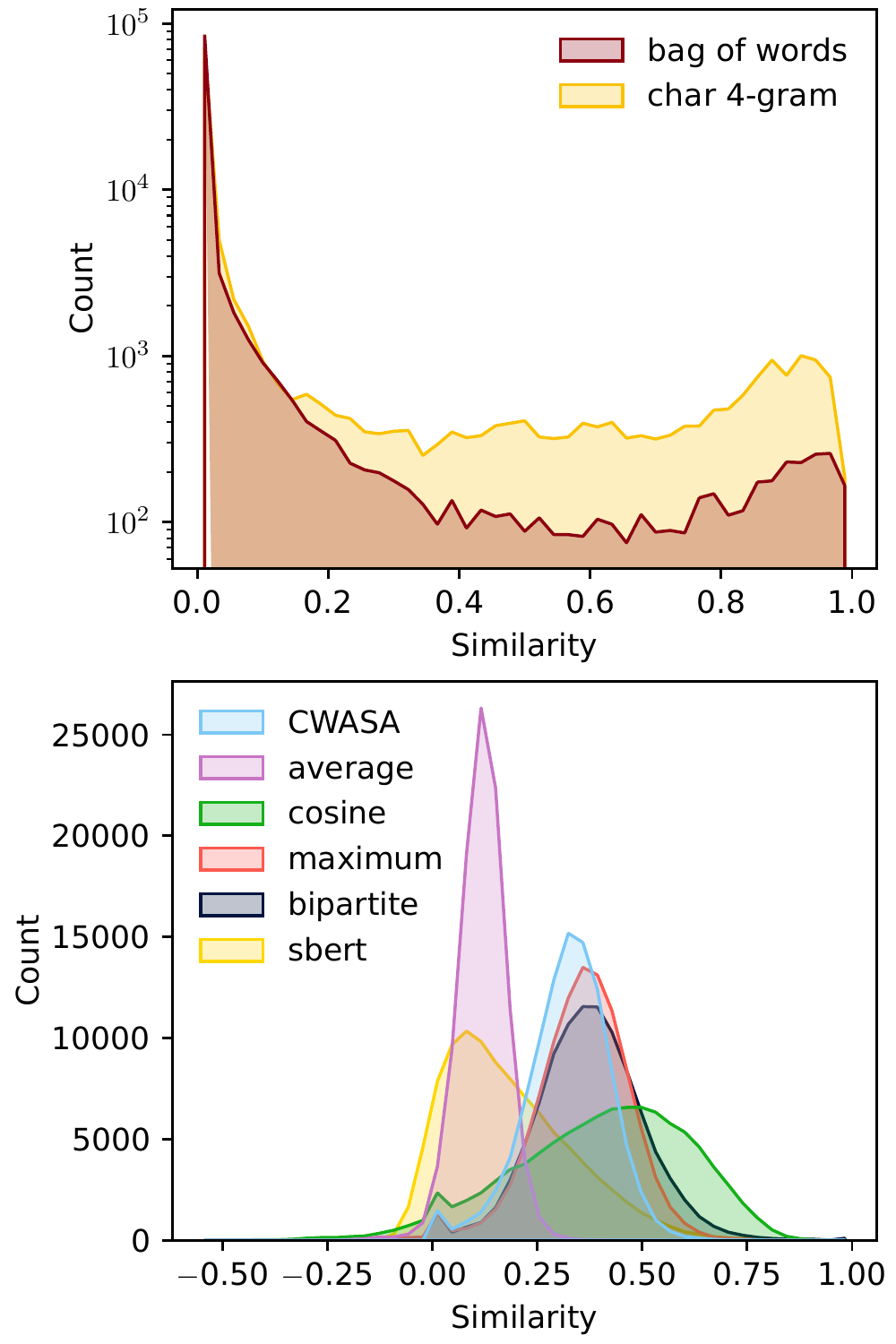}
    \caption{Histograms showing the distributions of the different similarity measures (top: TF-IDF, bottom: embedding based) evaluated on 100\,000 sentence pairs.}
    \label{fig:distributions}
\end{figure}

\begin{figure}[t]
    \centering
    \includegraphics[width=0.9\columnwidth]{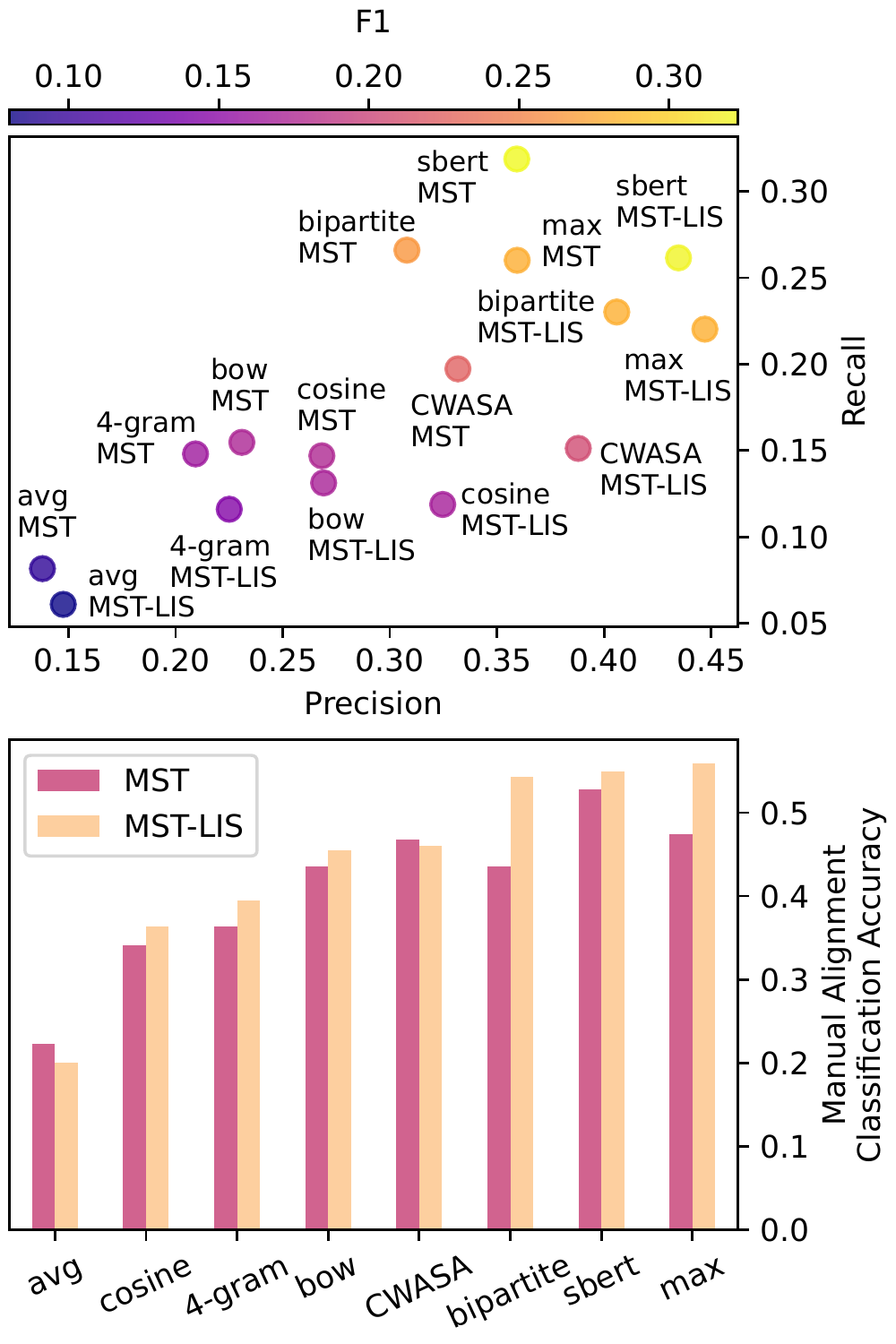}
    \caption{(top) Precision, recall and F1-score for all algorithm variants evaluated on the ground truth.\\ (bottom) Manual alignment classification accuracy for the manually labelled matches.}
    \label{fig:eval}
\end{figure}

We combine both matching algorithms with all eight similarity measures using either a threshold of $\mu + 1.5 \cdot \sigma$ or no threshold. This gives a total of 32 different alignment variants, the results of which we will discuss here.
We select the best algorithm variant according to a two stage process.
First, we analyse the results of the different alignment variants quantitatively.
Then, we perform two kinds of manual evaluation. For the first one we create a ground truth by manually aligning the sentences for a subset of articles. The second one focuses on the matches by manually labelling them as either correct or incorrect alignments.

\subsection{Quantitative Evaluation}
\label{sec:exploratory}

In \autoref{tab:quant-matches} we present -- for all algorithm variants -- the overall number of identified sentence matches. \autoref{tab:full-stats} adds information about their respective average similarity.
Depending on the choice of similarity measure, matching method, and threshold, between 7\,700 and 32\,500 matched sentences pairs are found in the entire corpus of a total of 32\,899 Simple German sentences. 
Introducing the threshold roughly halves the number of matches for the MST algorithm and results in only a third of matches for the MST-LIS algorithm if the similarity measure is kept fixed. 
Using MST yields more matches than using MST-LIS, which is expected as the latter is more restrictive. 
Quite surprisingly, the average similarity of the matches is only a little lower for MST than for the MST-LIS for any fixed choice of similarity measure and threshold value. 
Consequently, the average similarity allows no conclusions about the quality of the matches. 
Further, we notice that using the similarity threshold always results in a higher average similarity. 

\autoref{fig:distributions} gives an overview of the distributions of the similarity values over 100\,000 randomly sampled sentence pairs for all similarity measures.
The majority of the similarity values for the TF-IDF based methods 
is zero.
We plot the corresponding graph (top) with log-scale.
This observation is intuitive, as the value of these sentence similarity strategies is always zero if the two evaluated sentences do not have a word (or 4-gram) in common.
In contrast, the word embedding based methods (bottom) show a different distribution. 
Both, the average and SBERT similarity measure are unimodally distributed, the other similarity measures show one distinct peak and another small peak close to zero. 
However, the range of values and therefore the standard deviation seems to be particularly small for the average similarity measure.

\subsection{Manual Evaluation}
\label{sec:manual_evaluation}

For a first analysis, we create a ground truth of sentence alignments by manually labelling a subset of articles, sampling uniformly $39$ articles from the corpus.
This allows us to evaluate the alignment algorithms with respect to precision, recall, and F1-score. 
To this end, we built a simple GUI, see \autoref{fig:GUI-gt}, that presents the sentences of both articles side by side, allowing us to find the $n:1$ matches of Simple German and German sentences. 
We consider additional simple sentences explaining difficult concepts as part of the alignment, as long as they are a maximum of two sentences away from the literal translation of the German source sentence. 
We observe that depending on the source, the articles in Simple German are barely a translation of the original article. Besides, the order of information is often not maintained and in general, we only matched on average 33~\% of all German sentences.
\autoref{fig:eval} (top) shows the results for all 32 algorithm variants on the ground truth. SBERT, bipartite, and maximum similarity show good results. SBERT achieves the highest F1 score of $0.32$ with precision and recall at $0.43$ and $0.26$, respectively. While maximum similarity achieves a lower F1 score, its precision of $0.45$ is higher.

Complementary to the first analysis, we continue by focusing only on the matches of each alignment algorithm. For the manual evaluation of the alignment, we randomly sample 4\,627 sentence pairs from the set of aligned sentences obtained from all algorithm variants.
Given two sentences, it is inherently easier for a human annotator to make a yes/no decision whether the two presented sentences are a (partial) match or not. 
While this kind of evaluation does not allow any conclusions about the number of missed matches (i.e. recall) or the relation to additional explanatory sentences,  we argue that it gives a different perspective on the quality of the computed alignments as done by \citet{xu2015problems}.
As this analysis differs from the previous ground-truth set based analysis, we deliberately avoid the term precision and call the fraction of pairs that are labelled as (partial) matches as ``manual alignment classification accuracy''.
Thus, we created a different GUI, shown in \autoref{fig:GUI-matchings}, only displaying two sentences at a time and asking the annotator to label them as either ``match'' (likewise for partial matches) or ``no match''. The algorithm variant stays unknown to the user at evaluation time. 
\autoref{fig:eval} (bottom) shows the results of the manual alignment classification accuracy analysis. The ranks of the algorithm variants roughly correspond to the ranks under F1-score on the ground truth. Again, maximum similarity, SBERT, and bipartite similarity perform best. Maximum similarity with MST-LIS reaches the best manual alignment classification accuracy of 55.94~\%. \autoref{appendix:evaluation} presents detailed results and a per website analysis.

Finally, we create the sentence-level alignment using maximum similarity with MST-LIS, since it yields the highest precision on the ground truth and the highest manual alignment classification accuracy. \autoref{fig:examples} shows exemplary alignments.

\begin{table}
\centering
\begin{tabular}{llllr} 
    \toprule
    Matching & \multicolumn{2}{c}{MST} & \multicolumn{2}{c}{MST-LIS} \\
    \cmidrule(lr){2-3} \cmidrule(lr){4-5} 
    Threshold & \multicolumn{1}{c}{-} & \multicolumn{1}{c}{1.5} & \multicolumn{1}{c}{-} & \multicolumn{1}{c}{1.5}  \\ 
    \midrule
    
     bow           & 31\,733 & 18\,056 & 21\,026 & 10\,218 \\
     4-gram        & 32\,430 & 17\,861 & 27\,660 & 10\,649 \\
     cosine        & 32\,667 & 17\,684 & 31\,813 & 7\,781  \\
     average       & 29\,943 & 17\,257 & 29\,480 & 12\,575 \\
     CWASA         & 32\,696 & 19\,659 & 32\,516 & 9\,173  \\
     bipartite     & 32\,696 & 24\,142 & 32\,489 & 11\,854 \\
     maximum       & 32\,696 & 21\,506 & 32\,499 & 10\,304 \\
     sbert         & 32\,899 & 24\,314 & 32\,011 & 12\,982 \\

    \bottomrule
\end{tabular}
\caption{Number of matched sentences for all combinations of similarity measure and matching algorithm.}
\label{tab:quant-matches}
\end{table}
\section{Discussion}
\label{sec:discussion}
The results for the sentence alignments presented in Section~\ref{sec:alignment} show that the more sophisticated similarity measures perform better in terms of both F1-score and manual alignment classification accuracy.
The SBERT similarity is the most sophisticated similarity measure yielding the highest F1 score.
However, the precision and alignment classification accuracy of the maximum similarity with MST-LIS is higher. Generally, MST-LIS benefits from its strong assumption on the order of information in both articles yielding a higher accuracy, but in return not finding all possible alignments. This can be traced back to our observation, that Simple German articles often show a different structure.

\paragraph{Limitations}
Our work presents a new dataset based on text data scraped from the internet. Hence, the quality of the text depends on the quality of the available websites. 
Most of our data stems from the three websites \emph{apo}, \emph{koe} and \emph{mdr} providing a rich vocabulary in our corpus. While this vocabulary covers a variety of mixed topics, we cannot rule out any negative side effects of data imbalance. Moreover, our dataset can only represent topics that were considered relevant to be translated into Simple German by the respective website.

In Section~\ref{sec:manual_evaluation} we presented the different GUIs that we used to either manually align the sentence pairs or evaluate a sample of sentence alignments. One drawback of the tool for the second evaluation method is that it focuses solely on the matched sentences and presents them isolated from their contexts. One can argue that evaluators using the tool would have to see the context in which the sentences appear in order to correctly classify partial matches. Also, providing more information to the annotators might enable them to also correctly classify additional explanatory sentences.

\paragraph{Future Work and Use Cases}
Our corpus comprises data in LS and ES, two types of Simple German.
A higher granularity of language difficulty could be achieved by incorporating texts originally directed at language learners that are rated, e.g. according to the European Reference System \citep{cefr}.
Our work presents a parallel corpus for German and Simple German and should be continuously expanded. Not only to increase its size, but mainly to increase the number of topics covered in the corpus. Yet, as there are no efforts to start a single big corpus like a Simple German Wikipedia, web scraping from various sources stays the method of choice for the future.
An additional option is to compute sentence alignments for existing article aligned corpora  to include them in the dataset \citep[e.g.][]{battisti2020}.

As for the sentence alignment algorithms, various extensions are imaginable. Firstly, it might be interesting to allow one Simple German sentence to be matched to multiple German sentences. 
Also, the assumption of the MST-LIS about the order of information is very strong, and recall might be improved by softening this assumption, e.g. by allowing matches that are at most $n$ sentences away.
Other alignment algorithms that impose different biases on sentence order \citep{barzilay-elhadad-2003-sentence, jiang2020, Zhang2017} are interesting for further extensions.

Our dataset can be used to train (or fine tune) automatic text simplification systems \citep[e.g.][]{Xue2021mt5} which then should produce text with properties of Simple German.
Direct use cases for such simplification systems are support systems for human translators or \href{https://browser.mt/}{browser plugins} to simplify web pages. 
Further research has shown that text simplification as a pre-processing step may increase performance in downstream natural language processing tasks such as information extraction \citep{NiklausBHF16},  relation extraction \citep{van2021howcanihelp}, or machine translation \citep{StajnerP16simpletranslate}.
It remains an interesting direction for future research if Simple German can help to further increase performance on such tasks.
%
\section{Conclusion}
\label{sec:conclusion}

In this paper, we presented a new monolingual sentence-aligned extendable corpus for Simple German -- German that we make readily available. The data comprises eight different web sources and contains 708 aligned documents and a total of 10\,304 matched sentences using the maximum similarity measure and the MST-LIS matching algorithm. 
We have compared various similarity metrics and alignment methods from the literature and have introduced a variable similarity threshold that improves the sentence alignments.

We make the data accessible by releasing a URL collection\footnote{To preserve the dataset for as long as possible, we have archived all articles using the WayBackMachine by the internet archive.} as well as the accompanying code for creating the dataset, i.e. the code for the text pre-processing and sentence alignment.   
Our code can easily be adapted to create and analyze new sources. Even the application to non-German monolingual texts should be possible when specifying new word embeddings and adjusting the pre-processing steps.

We have obtained generally good results on our data. Our corpus is substantially bigger than the one in \citet{klaper2013} (708 compared to 256 parallel articles) and our results of the best sentence alignment methods are better as well (F1-scores: $0.28$ compared to $0.085$). It is also bigger than the parallel corpus created in \citet{battisti2020} (378 aligned documents), which does not provide any sentence level alignment.

\anonymize{}{
\section*{Acknowledgements}
This research has been funded by the Federal Ministry of Education and Research of Germany and the state of North Rhine-Westphalia as part of the Lamarr Institute for Machine Learning and Artificial Intelligence, LAMARR22B. 
Part of this work has been funded by the Vienna Science and Technology Fund (WWTF) project ICT22-059.
}

\bibliography{references}
\bibliographystyle{acl_natbib}

\clearpage

\appendix

\newpage
\appendix
\section{Datasheet}
\label{appendix:datasheet}

\subsection{Motivation for the Dataset Creation}
\question{For what purpose was the dataset created?}

\answer{Our dataset addresses the lack of a German dataset in simple language. During the creation of the dataset, we were primarily considering the problem of text simplification via neural machine translation. Hence, we worked to create a sentence-level alignment. Problems besides text simplification like automatic accessibility assessment, text summarization, and even curriculum learning would benefit from that data.}

\question{Who created the dataset (e.g. which team, research group) and on behalf of which entity (e.g. company, institution, organization)?}{}

\answer{\anonymize{The dataset was created by the authors.}{The dataset was created by the authors as part of the work of the MLAI Lab of the University of Bonn.}}

\question{Who funded the creation of the dataset?}

\answer{\anonymize{Funding information will be disclosed after double-blind review period has ended.}{This research has been funded by the Federal Ministry of Education and Research of Germany and the state of North Rhine-Westphalia as part of the Lamarr Institute for Machine Learning and Artificial Intelligence, LAMARR22B.
Part of this work has been funded by the Vienna Science and Technology Fund (WWTF) project ICT22-059.}}

\subsection{Composition}
\question{What do the instances that comprise the dataset represent (e.g. documents, photos, people, countries)?}

\answer{The instances comprise text from eight online resources organized per article per source. For each article in German, there exists an article in Simple German. We further publish the results of the proposed sentence-level alignment, where each German sentence has $n$ corresponding Simple German  sentences.}

\question{How many instances are there in total (of each type, if appropriate)?}{}

\answer{There are 712 articles (resp. 404\,771 tokens) in German and 708 articles (resp. 250\,093 tokens) in Simple German. For the sentence alignment there are 10\,304 matched sentences.}

\question{Does the dataset contain all possible instances or is it a sample (not necessarily random) of instances from a larger set?}

\answer{During the process of data collection we focused on German websites, we did not consider Swiss or Austrian resources. Further, the data collection was limited by the structure of the websites and the possibilities of the parser: some Simple German articles were excluded if they did not link to a corresponding version in  German. Also, some text sections might have been omitted due to the configuration of the \texttt{html}-blocks. No tests were run to determine the representativeness.}

\question{What data does each instance consist of?}

\answer{The parsed articles are structured by their respective source. Inside each source folder there is a json file with an entry per article containing all metadata consisting of the URL, the crawling date, the publishing date (if available), a flag whether the article from this URL is in simple language or not, a list of all associated articles, and the type of language (AS = Alltagssprache (everyday language), ES = Einfache Sprache (Simple German, less restrictive), LS = Leichte Sprache (Simple German, very restrictive)). 
Each article consists of text associated with one webpage. We removed \texttt{html} tags and performed light text pre-processing. 

Inside the results folder there exists an alignments folder with two files for each article. One file containing all aligned sentences in German and the other file containing the Simple German sentences at the corresponding line. Further, the results folder contains a json file recording the name of the original article and the similarity value for the two matched sentences according to the alignment method.}

\question{Is there a label or target associated with each instance?}

\answer{The instances do not have any labels, but each file of German text/sentences has a corresponding file with Simple German text/sentences.}

\question{Is any information missing from individual instances?}

\answer{As raised earlier, the websites were not crawled in their entirety, if there was no link provided from the Simple German to the German article. Also, text might have been omitted due to the limitations of the parser.}

\question{Are relationships between individual instances made explicit (e.g. users’ movie ratings, social network links)?}

\answer{There are no explicit relationships between individual instances recorded in our dataset, except for the alignments between Simple German articles and corresponding German articles. Any further links within articles were discarded during preprocessing.}

\question{Are there recommended data splits (e.g. training, development/ validation, testing)?}

\answer{No.}

\question{Are there any errors, sources of noise, or redundancies in the dataset?}

\answer{The dataset as a collection of textual data from different articles does not contain any errors. The quality of the sentence alignment is discussed in the paper.}

\question{Is the dataset self-contained, or does it link to or otherwise rely on external resources (e.g. websites, tweets, other datasets)?}

\answer{We publish the dataset as a URL collection. Instead of linking to the original articles, we archived the articles using the WayBackMachine by the internet archive. We provide the code to recreate the dataset.

Additionally, we provide a fully prepared version of the dataset upon request.}

\question{Does the dataset contain data that might be considered confidential (e.g. data that is protected by legal privilege or by doctor-patient confidentiality, data that includes the content of individuals non-public communications)?}

\answer{No.}

\question{Does the dataset contain data that, if viewed directly, might be offensive, insulting, threatening, or might otherwise cause anxiety?}

\answer{A majority of our data originates from a state-funded public broadcasting service. Thus, these texts may cover topics like criminal offenses, war, and crime. But we do not expect this to be the majority.}

\subsection{Collection Process}

\question{How was the data associated with each instance acquired?}

\answer{We crawled and processed directly observable textual data from eight different websites.}

\question{What mechanisms or procedures were used to collect the data (e.g. hardware apparatus or sensor, manual human curation, software program, software API)?}

\answer{We used the WayBackMachine\footnote{\url{https://archive.org/web/}} to archive the article URLs and the python library BeautifoulSoup\footnote{\url{ https://www.crummy.com/software/BeautifulSoup/}} to crawl the websites.}

\question{If the dataset is a sample from a larger set, what was the sampling strategy (e.g. deterministic, probabilistic with specific sampling probabilities)?}

\answer{We chose websites that offered parallel articles in German and Simple German, which were consistent in their linking between the articles.}

\question{Who was involved in the data collection process (e.g. students, crowdworkers, contractors) and how were they compensated (e.g. how much were crowdworkers paid)?}

\answer{All work for this dataset was done by persons that are listed among the authors of this paper. Part of this work has been done as a study project for which the students were given credit.}

\question{Over what timeframe was the data collected? Does this timeframe match the creation timeframe of the data associated with the instances (e.g. recent crawl of old news articles)?}

\answer{The data was collected over a timeframe of three months, November 2021 until January 2022. This does not necessarily correspond with the publication date of the articles.}

\question{Were any ethical review processes conducted (e.g. by an institutional review board)?}

\answer{No.}

\subsection{Preprocessing/ cleaning/ labeling}

\question{Was any preprocessing/ cleaning/ labeling of the data done (e.g. discretization or bucketing, tokenization, part-of-speech tagging, SIFT feature extraction, removal of instances, processing of missing values)?}

\answer{With the parsing of the websites light pre-processing was performed. We ignored images, \texttt{html}-tags, and corresponding text metadata. Also, enumerations were transformed into comma-separated text.}

\question{Was the “raw” data saved in addition to the preprocessed/ cleaned/ labeled data (e.g. to support unanticipated future uses)?}

\answer{By using the URLs to the archived, original articles, the raw data is part of this work.}

\question{Is the software used to preprocess/ clean/ label the instances available?}

\answer{All libraries and code are available at the time of publication.}

\subsection{Uses}
\question{Has the dataset been used for any tasks already?}

\answer{No.}

\question{Is there a repository that links to any or all papers or systems that use the dataset?}

\answer{This information will be stored in the repository on GitHub\footnote{\label{github}\anonymize{Anonymized.}{\url{https://github.com/mlai-bonn/Simple-German-Corpus}}}.}

\question{What (other) tasks could the dataset be used for?}

\answer{Language modelling and monolingual neural machine translation for text simplification, text accessibility, possibly also latent space disentanglement or as a baseline for what constitutes simple language.}

\question{Is there anything about the composition of the dataset or the way it was collected and preprocessed/ cleaned/ labeled that might impact future uses?}

\answer{The original sources are archived and should remain publicly available, allowing novel use cases that we did not foresee.}

\question{Are there tasks for which the dataset should not be used?}

\answer{This dataset is composed of eight online resources that are either about social services, German news, general health information, or include administrative information. The potential limitations of the vocabulary of this corpus should be considered before training systems with it.}

\subsection{Distribution}
\question{Will the dataset be distributed to third parties outside of the entity (e.g. company, institution, organization) on behalf of which the dataset was created?}

\answer{Yes, the dataset will be publicly available. Due to legal concerns, we make publicly available:
\begin{itemize}
    \item A list of URLs to parallel articles that were archived in the Wayback machine of the Internet archive
    \item code to download the articles and do all processing steps described in this article, using the list of URLs.
\end{itemize}
We share a readily available dataset upon request.}

\question{How will the dataset be distributed (e.g. tarball on website, API, GitHub)}

\answer{The dataset will be distributed via GitHub\textsuperscript{\ref{github}}.}

\question{When will the dataset be distributed?}

\answer{The dataset was released in 2022.}

\question{Will the dataset be distributed under a copyright or other intellectual property (IP) license, and/or under applicable terms of use (ToU)?}

{We publish the dataset under the CC BY-SA 4.0 license as a URL collection and the accompanying code to easily recreate the dataset under MIT license. 
In order to ensure the long-term availability of the sources, we archived them in the \href{https://archive.org/web/}{Internet Archive}.
We further share the entire, ready-to-use dataset upon request via email.}

\question{Have any third parties imposed IP-based or other restrictions on the data associated with the instances?}

\answer{No.}

\question{Do any export controls or other regulatory restrictions apply to the dataset or to individual instances?}

\answer{No.}

\subsection{Maintenance}
\question{Who will be supporting/ hosting/ maintaining the dataset?}

\answer{The dataset will be maintained via the GitHub repository.}

\question{How can the owner/ curator/ manager of the dataset be contacted (e.g. email address)?}

\answer{The creators of the dataset can be contacted via GitHub and e-mail: \anonymize{(anonymized)}{\href{mailto:toborek@cs.uni-bonn.de}{toborek@cs.uni-bonn.de}}.}

\question{Is there an erratum?}

\answer{Not at the time of the initial release. However, we plan to use GitHub issue tracking to work on and archive any errata.}

\question{Will the dataset be updated (e.g. to correct labeling errors, add new instances, delete instances)?}

\answer{Updates will be communicated via GitHub. We plan to extend the work in the future, by adding new articles. Deletion of individual article pairs is not planned at the moment.}

\question{If the dataset relates to people, are there applicable limits on the retention of the data associated with the instances (e.g. were individuals in question told that their data would be retained for a fixed period of time and then deleted)?}

\answer{Not applicable.}

\question{Will older versions of the dataset continue to be supported/hosted/maintained?}

\answer{All updates will be communicated via GitHub. Versioning will be done using git tags, which ensures that previously released versions of the dataset and code base will stay available.}

\question{If others want to extend/ augment/ build on/ contribute to the dataset, is there a mechanism for them to do so?}

\answer{We hope that others will contribute to the dataset in order to improve the dataset landscape for German language.  The code is modular and we invite the community to add new instances, websites, and corresponding crawlers as well as alignment strategies and similarity measures. We invite collaboration via personal communication and/or GitHub pull requests.}


\section{Dataset Description}

We have created a corpus consisting of 708 Simple German  and 712 corresponding German articles from eight web sources spanning different topics. Few Simple German articles are matched to multiple German ones, and the other way around.
\autoref{tab:sources} shows the eight different online websites and gives an overview of each website's content. After using the proposed algorithm variants of maximum similarity with MST-LIS matching and a similarity threshold of $1.5$, we obtain a total of 10\,304 sentence pairs. In \autoref{tab:sentence-alignments} we consider in detail the number of $n:1$ aligned sentence pairs originating from each website. 

\begin{table*}
    \centering
    \resizebox{\textwidth}{!}{
    \begin{tabular}{llccc}
        \toprule
         Website & Content & Simple & Standard & Type\\
        \midrule
        (apo) \href{https://www.apotheken-umschau.de/einfache-sprache/}{apotheken-umschau.de} & General health information & 168 & 166 & ES\\
        (beb) \href{https://www.behindertenbeauftragter.de/DE/AS/startseite/startseite-node.html}{behindertenbeauftragter.de}  & Official office for disabled people & 21 & 21 & LS\\
        (bra) \href{https://www.brandeins.de/themen/rubriken/leichte-sprache}{brandeins.de}  & Translating excerpts from various topics & 47 & 47 & LS\\
        (lmt) \href{https://www.lebenshilfe-main-taunus.de/}{lebenshilfe-main-taunus.de}  & Non-profit association for disabled people & 45 & 45 & LS\\
        (mdr) \href{https://www.mdr.de/nachrichten-leicht/index.html}{mdr.de} & State-funded public broadcasting service & 322 & 322 & LS\\
        (soz) \href{https://www.sozialpolitik.com/}{sozialpolitik.com} & Explains social policy in Germany & 15 & 15 & LS\\
        (koe) \href{https://www.stadt-koeln.de/leben-in-koeln/soziales/informationen-leichter-sprache}{stadt-koeln.de} & Administrative information (City of Cologne) & 82 & 82 & LS\\
        (taz) \href{https://taz.de/Politik/Deutschland/Leichte-Sprache/!p5097/}{taz.de}  & German Newspaper (discontinued) & 8 & 14 & LS\\
        \addlinespace
        Total &   & 708 & 712 \\
        \bottomrule
    \end{tabular}
    }
    \caption{Overview of websites used for the corpus. One website \emph{(apo)} offers general health information, three websites \emph{(bra)}, \emph{(mdr)} and \emph{(taz)} are news websites, three websites \emph{(beb)}, \emph{(lmt)} and \emph{(soz)} offer information about different social services, and \emph{(koe)} provides administrative information about the city of Cologne. The last column describes the type of Simple German found on that website `Einfache Sprache' (ES) or `Leichte Sprache' (LS).}
    \label{tab:sources}
\end{table*}

\begin{table*}
        \begin{tabular}{llllllllll}
            \toprule
                      & (apo)      & (beb)  & (bra)  & (lmt)  & (mdr)   & (soz)   & (koe)   & (taz)  & Total    \\
            \midrule
            Sentences & 2\,311     & 223    & 195    & 275    & 1\,505  & 180     & 1\,132  & 121    & 5\,942  \\
            Tokens    & 33\,847    & 3\,885 & 3\,896 & 3\,370 & 23\,348 & 2\,523  & 16\,561 & 1\,981 & 89\,411 \\
            \bottomrule
        \end{tabular}
    \centering
    \caption{Overview of the number of aligned sentence pairs yielded by our proposed algorithm variant of maximum similarity with MST-LIS matching and a similarity threshold of $1.5$, where we count the number of unique sentences in German with $n$ corresponding Simple German sentences.}
    \label{tab:sentence-alignments}
\end{table*}

\section{Similarity Measures}
\label{appendix:ext_similarity_measures}

We describe an article $A$ as a list of sentences, i.e. $A = \mylist{s_1, \ldots, s_n}$. We define $A^S$ and $A^C$ as the simple and complex versions of the same article with $\abs{A^S} = n$ and $\abs{A^C} = m$. We consider a variant of the sentence alignment problem that receives two lists of sentences $A^S$ and $A^C$ and produces a list of pairs $\mylist{(s_i^S, s_j^C)}$ such that, with relative certainty, $s_i^S$ is a (partial) simple version of the complex sentence $s_j^C$. 
Given two lists of pre-processed sentences $A^S$ and $A^C$, we compute similarities between any two sentences $s_i^S \in A^S$ and $s_j^C \in A^C$. A sentence can be described either as a list of words $s_i^S = \mylist{w_1^S, \ldots, w_l^S}$ or as a list of characters $s_i^S = \mylist{c_1^S, \ldots, c_k^S}$. 
In total, we have compared eight different similarity measures. Two of the measures are based on TF-IDF, the other six rely on word embeddings. We have decided to use the pre-trained word embeddings supplied by spaCy in the \texttt{d\_core\_news\_lg}\footnote{\url{https://spacy.io/models/de\#de_core_news_lg}} bundle and the pre-trained \texttt{distiluse-base-multilingual-cased- v1} model provided by \citet{reimers2019}. 
\autoref{tab:full-stats} shows average similarity values of matching sentences and number of resulting matches for all combinations of similarity measures and alignment strategies.

\begin{table*}
\begin{center}
    \begin{tabular}{llllllllr} 
    \toprule
    \multicolumn{1}{l}{} & \multicolumn{4}{c}{Average Similarity} & \multicolumn{4}{c}{Sentence Matches} \\ 
    \cmidrule(lr){2-5} \cmidrule(lr){6-9}
    Alignment & \multicolumn{2}{c}{MST} & \multicolumn{2}{c}{MST-LIS} & \multicolumn{2}{c}{MST} & \multicolumn{2}{c}{MST-LIS} \\
    \cmidrule(lr){2-3} \cmidrule(lr){4-5} \cmidrule(lr){6-7} \cmidrule(lr){8-9}
    Threshold & \multicolumn{1}{c}{-} & \multicolumn{1}{c}{1.5} & \multicolumn{1}{c}{-} & \multicolumn{1}{c}{1.5} & \multicolumn{1}{c}{-} & \multicolumn{1}{c}{1.5} & \multicolumn{1}{c}{-} & \multicolumn{1}{c}{1.5}  \\ 
    \midrule
    
     bag of words & 0.37 & 0.59 & 0.32 & 0.60 & 31\,733 & 18\,056 & 21\,026 & 10\,218  \\
     4-gram       & 0.45 & 0.68 & 0.35 & 0.68 & 32\,430 & 17\,861 & 27\,660 & 10\,649  \\
     cosine       & 0.68 & 0.77 & 0.57 & 0.77 & 32\,667 & 17\,684 & 31\,813 & 7\,781    \\
     average      & 0.17 & 0.21 & 0.15 & 0.20 & 29\,943 & 17\,257 & 29\,480 & 12\,575    \\
     CWASA        & 0.49 & 0.56 & 0.43 & 0.56 & 32\,696 & 19\,659 & 32\,516 & 9\,173   \\
     bipartite    & 0.65 & 0.71 & 0.56 & 0.71 & 32\,696 & 24\,142 & 32\,489 & 11\,854   \\
     maximum      & 0.56 & 0.62 & 0.49 & 0.63 & 32\,696 & 21\,506 & 32\,499 & 10\,304   \\
     sbert        & 0.49 & 0.57 & 0.38 & 0.57 & 32\,899 & 24\,314 & 32\,011 & 12\,982  \\
    
    \bottomrule
\end{tabular}
\end{center}
\caption{Average similarity values and number of matched sentences for all combinations of similarity measure and alignment strategy.}
\label{tab:full-stats}
\end{table*}

\paragraph{Bag of words similarity} Following \citet{paetzold2017}, we calculate for each $w \in s_i$ the TF-IDF values. 
TF-IDF is the product of the term frequency (TF) \citep{Luhn1957} and the inverse document frequency (IDF) \citep{sparckjones1972} given a word and its corpus.
We then weigh each sentence's bag of words vector by its TF-IDF vector before calculating the cosine similarity between them:




\begin{equation}
    \begin{aligned}
        \label{eqn:tf-idf}
        &\sentsim(s^S, s^C) = \\
        &\frac{\sum\limits_{w \in s^S \cap s^C}\tfidf(w^S) \cdot \tfidf(w^C)}{\sqrt{\sum\limits_{w^S \in s^S} \tfidf(w)^2 \cdot \sum\limits_{w^C \in s^C} \tfidf(w)^2}} \; .   
    \end{aligned}
\end{equation}

\paragraph{Character 4-gram similarity} This method works analogously to the TF-IDF method, but instead of taking into account the words, it uses character n-grams, which span the word boundaries.
We have decided to follow the results from \cite{mcnamee2004}, who have determined $n=4$ to be performing best for German text.

\paragraph{Cosine similarity}
We use pre-calculated word embeddings to calculate the cosine similarity using the average of each sentence's word vectors \citep{stajner2018, mikolov2013}. 
Let $\emb(w)$ be the embedding vector of word $w$ 
and let $\cossim(\Vec{v}, \Vec{w}) = \frac{\Vec{v} \, \cdot \, \Vec{w}}{\norm{v}\norm{w}}$ be the cosine similarity between two vectors, 
then the vector similarity is
\begin{equation}
    \begin{aligned}
    \label{eqn:cosine}
        &\sentsim(s^S, s^C) = \\
        &\cossim\left(\sum_{w^S \in s^S} \emb\left(w^S\right), \sum_{w^C \in s^C} \emb\left(w^C\right)\right) \;.
    \end{aligned}
\end{equation}

\paragraph{Average similarity}
For all pairs of words in a given pair of $(A^S, A^C)$ \citep{kajiwara2016} we use the embedding vector $\emb(w)$ of each word $w$ to calculate the cosine similarity $\cossim$ between them. The average similarity is defined as following, where $\phi(w^S, w^C) = \cossim(\emb(w^S), \emb(w^C))$:

\begin{equation}
\begin{aligned}
    &\operatorname{AvgSim}(s^S, s^C) = \\ 
    &\frac{1}{|s^S|\cdot|s^C|} \sum_{w^S \in s^S} \sum_{w^C \in s^C} \phi(w^S, w^C).
\end{aligned}
\end{equation}

\paragraph{CWASA} The Continuous Word Alignment-based Similarity Analysis method was presented by \citet{franco-salvador2015} and implemented by \citet{stajner2018}.
Contrary to the previous similarity measure, it does not average the embedding vector values. Instead, it finds the best matches for each word in $s^S$ and in $s^C$ with $\cossim \geq 0$. Let $M^S = \{(w_1^S, w_i^C), \ldots, (w_l^S, w_j^C)\}$ be the set of best matches for the simple words, and $M^C=\{(w_i^S, w_1^C), \ldots, (w_j^S, w_m^C)\}$ be the set of best matches for the complex words. Then, 

\begin{equation}
\begin{aligned}
    &\operatorname{CWASA}(s^S, s^C) = \\
    &\frac{1}{\abs{M^S \cup M^C}} \sum_{(w^S, w^C) \in M^S \cup M^C} \phi(w^S, w^C).
\end{aligned}
\end{equation}

\paragraph{Maximum similarity} Similar to CWASA, \citet{kajiwara2016} calculate optimal matches for the words in both sentences.
The difference is that instead of taking the average of all word similarities $\geq 0$, only the maximum similarity for each word in a sentence is considered.
Let the asymmetrical maximal match be $\asym^S(s^S, s^C) = \frac{1}{\abs{M^S}} \sum_{(w_i^S, w_j^C) \in M^S} 
\cossim(\emb(w_i^S), \emb(w_j^C))$  
(and $\asym^C$ analogously), then
\begin{equation}
\label{eqn:max-sim}
\begin{aligned}
    &\operatorname{MaxSim}(s^S, s^C) = \\ 
    &\frac{1}{2}(\asym^S(s^S, s^C) + \asym^C(s^S, s^C)) \; .
\end{aligned}
\end{equation}

\paragraph{Bipartite similarity}
This method calculates a maximum matching on the weighted bipartite graph induced by the lists of simple and complex words  \cite{kajiwara2016}. Edges between word pairs are weighted with the word-to-word cosine similarity. The method returns the average value of the edge weights in the maximum matching.
The size of the maximum matching is bounded by the size of the smaller sentence.

\paragraph{SBERT similarity}
This method works similarly to the cosine similarity, but instead of using pre-calculated word embeddings, we use a pre-trained, multilingual Sentence-BERT \cite{reimers2020} to create contextualized embeddings for the entire sentence:
\begin{equation}
    \operatorname{SBERT}(s^S, s^C) = \\
    \cossim(\emb(s^S), \emb(s^C))
\end{equation}

\section{Evaluation}
\label{appendix:evaluation}

\begin{figure*}
    \centering
    \includegraphics[scale=0.4]{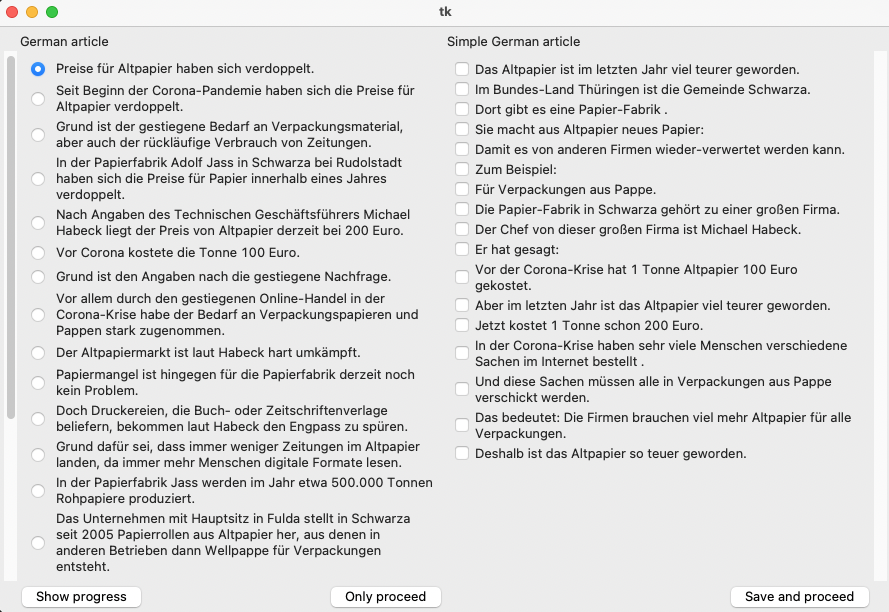}
    \caption{Simple GUI used to create the ground truth of sentence alignments.}
    \label{fig:GUI-gt}
\end{figure*}

\begin{figure*}
    \centering
    \begin{subfigure}[b]{0.6\textwidth}
     \centering
     \includegraphics[width=\textwidth]{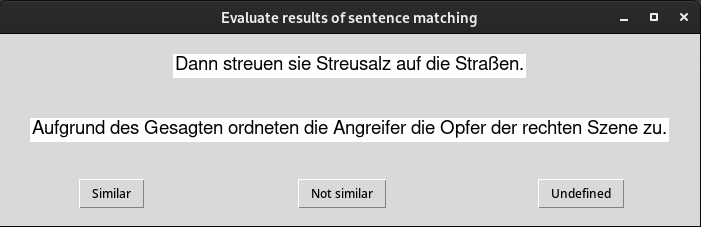}
     \caption{Example for an incorrect match (top) Simple German: ``Then they spread de-icing salt on the roads.'' and (bottom) German: ``Based on what was said, the attackers associated the victims with the right-wing scene.''}
     \label{fig:gui_not_similar}
     \end{subfigure}
     \hfill
     \begin{subfigure}[b]{0.6\textwidth}
     \centering
     \includegraphics[width=\textwidth]{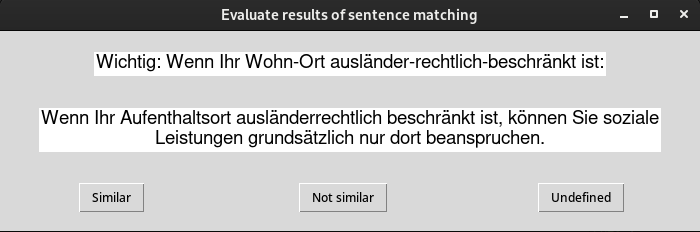}
     \caption{Example for a partial match (top) Simple German: ``Important: if your place of residence is restricted by immigration law.'' (bottom) German: ``If your place of residence is restricted under immigration law, you can generally only claim social benefits there.''}
     \label{fig:gui_partial}
     \end{subfigure}
     \hfill
     \begin{subfigure}[b]{0.6\textwidth}
     \centering
     \includegraphics[width=\textwidth]{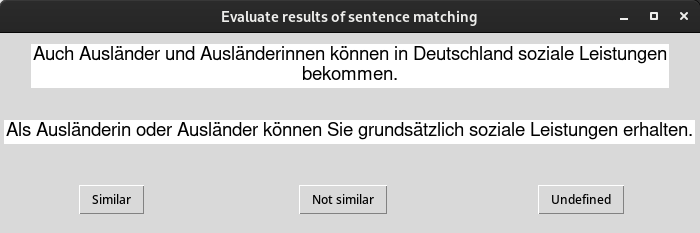}
     \caption{Example for a good match (top) Simple German: ``Foreigners can also receive social benefits in Germany.'' (bottom) German: ``As a foreigner, you are generally eligible for social benefits.''}
     \label{fig:gui_similar}
     \end{subfigure}
     \caption{GUI for the second manual evaluation. We show different examples of matches, and how they are evaluated.}
     \label{fig:GUI-matchings}
\end{figure*}

We performed two kinds of manual evaluation. For the first one, we created a ground truth by manually aligning the sentences for a subset of articles. Here, we report precision, recall, and F1-score based on the ground truth. The second evaluation focuses on the matches that are computed by our alignment methods by manually labelling them as either correct or incorrect. Here, we report alignment classification accuracy.
In \autoref{tab:all-f1-mst-list} we show the results of the ground-truth evaluation, broken down for each website. We can clearly see that the quality of the sentence alignment highly depends on the source. 
Further, in \autoref{fig:GUI-gt} we show the GUI that we used to create the ground truth of sentence alignments for a subset of articles.
\autoref{tab:accuracies} shows the exact precision values for the second manual evaluation that only considered the matches produced by each algorithm variant. 
Equally, in \autoref{fig:GUI-matchings} we show the different GUI for the evaluation of the matches.

\begin{table*}
    \centering
    \scalebox{0.90}{
        \begin{tabular}{llcccccccc}
    \toprule
    \begin{tabular}[c]{@{}l@{}}Similarity \\ Measure\end{tabular}  &           & bag of words & 4-gram & cosine & average & CWASA  & maximum & bipartite & sbert\\
    \cmidrule{1-1}
    Website            &           &              &        &        &         &        &         &         &         \\
    \midrule
    (apo)              & Precision & 0.30     & 0.20   & 0.26   & 0.11    & 0.33   & 0.43   & 0.41     & 0.43  \\
                       & Recall    & 0.23     & 0.13   & 0.10   & 0.05    & 0.24   & 0.36   & 0.35     & 0.39  \\
                       & F1        & 0.25     & 0.15   & 0.14   & 0.06    & 0.26   & 0.38   & 0.37     & 0.40  \\
    \midrule
    (beb)              & Precision & 0.78     & 0.82   & 0.82   & 0.75    & 0.91   & 0.92    & 0.70    & 0.88 \\
                       & Recall    & 0.41     & 0.40   & 0.33   & 0.13    & 0.38   & 0.52    & 0.42    & 0.60 \\
                       & F1        & 0.53     & 0.53   & 0.46   & 0.22    & 0.53   & 0.66    & 0.52    & 0.71 \\
    \midrule
    (bra)              & Precision & 0.47     & 0.43   & 0.34   & 0.02    & 0.56   & 0.76    & 0.70    & 0.56 \\
                       & Recall    & 0.15     & 0.13   & 0.07   & 0.01    & 0.10   & 0.19    & 0.20    & 0.26 \\
                       & F1        & 0.22     & 0.20   & 0.10   & 0.01    & 0.16   & 0.30    & 0.30    & 0.36 \\
    \midrule
    (lmt)              & Precision & 0.54     & 0.45   & 0.76   & 0.43    & 0.58   & 0.64    & 0.66    & 0.61 \\
                       & Recall    & 0.26     & 0.20   & 0.30   & 0.18    & 0.25   & 0.34    & 0.37    & 0.40 \\
                       & F1        & 0.35     & 0.28   & 0.42   & 0.25    & 0.35   & 0.44    & 0.46    & 0.48 \\
    \midrule
    (mdr)              & Precision & 0.22     & 0.15   & 0.27   & 0.10    & 0.37   & 0.38    & 0.31    & 0.40 \\
                       & Recall    & 0.10     & 0.09   & 0.09   & 0.05    & 0.12   & 0.15    & 0.16    & 0.20 \\
                       & F1        & 0.13     & 0.11   & 0.13   & 0.06    & 0.18   & 0.21    & 0.21    & 0.26 \\
    \midrule
    (soz)              & Precision & 0.05     & 0.00   & 0.32   & 0.26    & 0.29   & 0.29    & 0.33    & 0.36 \\
                       & Recall    & 0.03     & 0.00   & 0.17   & 0.19    & 0.17   & 0.21    & 0.23    & 0.26 \\
                       & F1        & 0.04     & 0.00   & 0.22   & 0.22    & 0.21   & 0.25    & 0.27    & 0.30 \\
    \midrule
    (koe)              & Precision & 0.07     & 0.12   & 0.19   & 0.04    & 0.23   & 0.32    & 0.35    & 0.32 \\
                       & Recall    & 0.03     & 0.11   & 0.10   & 0.03    & 0.10   & 0.21    & 0.22    & 0.18 \\
                       & F1        & 0.04     & 0.10   & 0.13   & 0.04    & 0.13   & 0.24    & 0.26    & 0.23 \\
    \midrule
    (taz)              & Precision & 0.00     & 0.00   & 0.04   & 0.00    & 0.04   & 0.06    & 0.11    & 0.10 \\
                       & Recall    & 0.00     & 0.00   & 0.03   & 0.00    & 0.03   & 0.06    & 0.18    & 0.12 \\
                       & F1        & 0.00     & 0.00   & 0.04   & 0.00    & 0.04   & 0.06    & 0.13    & 0.11 \\
    \midrule
    Average            & Precision & 0.27     & 0.23   & 0.32   & 0.15    & 0.39   & 0.45    & 0.41    & 0.43 \\
                       & Recall    & 0.13     & 0.12   & 0.12   & 0.06    & 0.15   & 0.22    & 0.23    & 0.26 \\
                       & F1        & 0.17     & 0.15   & 0.17   & 0.12    & 0.21   & 0.28    & 0.28    & 0.32 \\
    \bottomrule
\end{tabular}
    }
    \caption{Precision, recall, and F1-score results from the first evaluation on the ground truth per website. We compare the results of each similarity measure applied with the MST-LIS matching algorithm and a similarity threshold of $1.5$.}
    \label{tab:all-f1-mst-list}
\end{table*}

\begin{table*}
    \centering
    \begin{tabular}{lcc}
        \toprule
         & \multicolumn{2}{c}{Alignment classification accuracy} \\
        \cmidrule{2-3}
        Matching Strategy & \multicolumn{1}{c}{MST} & \multicolumn{1}{c}{MST-LIS} \\
        \midrule
        4-gram       & 0.36 & 0.39 \\
        CWASA        & 0.47 & 0.46 \\
        average      & 0.22 & 0.20 \\
        bag of words & 0.44 & 0.45 \\
        bipartite    & 0.43 & 0.54 \\
        cosine       & 0.34 & 0.36 \\
        maximum      & 0.47 & 0.56 \\
        sbert        & 0.53 & 0.55 \\
        \bottomrule
    \end{tabular}
    \caption{Alignment classification accuracy results from the second manual evaluation. All algorithm variants were tested with a threshold of $1.5$. Given two sentences, the annotators evaluate whether the sentence in Simple German is a (partial) translation of the German sentence.}
    \label{tab:accuracies}
\end{table*}


\end{document}